\documentclass[lettersize,journal]{IEEEtran}
\usepackage{amsmath,amsfonts}
\usepackage{algorithmic}
\usepackage{algorithm}
\usepackage{array}
\usepackage[caption=false,font=normalsize,labelfont=sf,textfont=sf]{subfig}
\usepackage{textcomp}
\usepackage{stfloats}
\usepackage{url}
\usepackage{verbatim}
\usepackage{graphicx}
\usepackage{xcolor}
\usepackage{booktabs,makecell}
\usepackage{multirow}
\usepackage{cite}
\begin{document}

\title{Confidence-Gated Vision-Only Heading Alignment for UAV–UGV Cooperative Systems}

\author{Reza Ahmari,~\IEEEmembership{Graduate Student member,~IEEE,}, Vahid Hemmati, Parham Kebria,~\IEEEmembership{Senior member,~IEEE,}, Olusola Odeyomi, Kaushik Roy, and Abdollah Homaifar
\thanks{R. Ahmari, O. Odeyomi, and K. Roy are with the Department of Computer Science, North Carolina Agricultural and Technical State University, Greensboro, NC 27411, USA. E-mail: rahmari@aggies.ncat.edu, otodeyomi@ncat.edu, kroy@ncat.edu.} \thanks{V. Hemmati, P. Kebria, and A. Homaifar are with the Department of Electrical and Computer Engineering, North Carolina Agricultural and Technical State University, Greensboro, NC 27411, USA. E-mail: vhemmati@ncat.edu, pmkebria@ncat.edu, homaifar@ncat.edu.} \thanks{Corresponding author: Abdollah Homaifar.}
}



\maketitle
\begin{center} \small This work has been submitted to the IEEE for possible publication. Copyright may be transferred without notice, after which this version may no longer be accessible. \end{center}
\begin{abstract}
Vision-based heading prediction is useful for UAV--UGV cooperation, but accurate prediction alone does not guarantee that every predicted heading should be issued directly as a control command. This paper investigates the decision problem of when and how a fixed vision-based heading predictor should be trusted for command issuance. A lightweight confidence-gated framework is proposed in which execution decisions are made using two interpretable reliability proxies derived from the perception stream: bounding-box area as a visibility-related proxy and short-window variation in predicted heading as a stability-related proxy. During low-confidence intervals, the framework compares the baseline freeze-HOLD policy with a bounded-blend fallback that updates the issued command conservatively.

The method is evaluated on a real UAV--UGV dataset under clean and perturbed conditions. The results show that confidence gating creates a clear trade-off among execution rate, executed-frame accuracy, issued-command accuracy, and smoothness. The results further show that sparse execution can cause severe stale-command error under the baseline freeze-HOLD policy, whereas the bounded-blend fallback substantially improves command-level behavior under the same gate decisions. These findings highlight that reliable perception-driven autonomy depends not only on prediction accuracy, but also on decision-aware command issuance during low-confidence intervals.
\end{abstract}


\section{INTRODUCTION}

Cooperative aerial--ground robotic systems have attracted sustained attention because they combine complementary sensing, mobility, and operational capabilities. In particular, unmanned aerial vehicles (UAVs) provide rapid viewpoint adaptation and broad situational awareness, while unmanned ground vehicles (UGVs) offer persistent ground-level interaction, payload support, and endurance. Recent literature shows that UAV--UGV collaboration is being actively studied across applications such as environmental monitoring, disaster response, inspection, and precision agriculture, while also highlighting persistent challenges related to sensing reliability, communication, coordination, and environmental uncertainty \cite{munasinghe2024uavugv, liu2022airground, ren2024agriculture, shahar2025ugvuav, chen2025multirobot, vaidhun2021dynamic, chen2024evolutionary, ahmari2025data}.

Within this broader setting, reliable relative perception between aerial and ground platforms is a critical requirement. Vision-based methods are especially attractive because they can support following, rendezvous, and landing behaviors without relying exclusively on external infrastructure. Prior work has demonstrated that UAVs can autonomously follow and land on moving ground platforms using visual sensing, including approaches based on fiducial landmarks, monocular perception, visual servoing, onboard sensing and computing, and more recent marker-less formulations \cite{morales2023vision, falanga2017vision, wu2019autonomous, araar2017landing, feng2018landing, keipour2022visualservoing, cui2023coarse, liu2025uwbvision}. These studies show that vision can provide the geometric information needed for relative pose or heading estimation in dynamic UAV--UGV interaction scenarios.

In a prior work, a vision-based heading predictor was developed to estimate relative heading from image-derived geometric features \cite{ahmari2025visual}. The present paper does not redesign or retrain such a predictor. Instead, it addresses a different and practically important question: \emph{when and how should a predicted heading actually be issued as a control command?} A perception model may achieve low average prediction error under nominal conditions, yet still produce unstable or unreliable outputs under degraded visibility, detection jitter, or shifted operating conditions. This distinction between prediction quality and decision reliability is well aligned with the broader uncertainty literature in machine learning and computer vision \cite{gal2016dropout}. Existing work has shown the importance of distinguishing aleatoric and epistemic uncertainty, obtaining practical uncertainty estimates in deep models, calibrating neural-network confidence, and evaluating model trustworthiness under dataset shift \cite{kendall2017uncertainties, gal2016dropout, lakshminarayanan2017deep, guo2017calibration, hendrycks2017ood, ovadia2019trust, schwaiger2020uncertainty, araujo2024roadsafety}. Related work on selective prediction further reinforces the principle that abstention or deferred action may be preferable when confidence is insufficient \cite{geifman2017selective, geifman2019selectivenet, yildirim2019selective}.

This issue is especially important in heading alignment and landing-related control loops \cite{ahmari2025smc, ahmari2025evaluating}. If every predicted heading is executed immediately, short-term perception variability can translate into oscillatory commands, unnecessary corrections, or unreliable actuation under degraded sensing conditions \cite{michelmore2018uqdriving, zhang2014failureprediction, araujo2024roadsafety}. Conversely, if the system simply withholds uncertain predictions and indefinitely freezes the previously issued command, prolonged low-confidence intervals can lead to stale-command accumulation. Therefore, the central problem addressed in this paper is not how to construct a heading predictor from scratch, but how to determine whether the current predicted heading from an existing perception module is sufficiently reliable for execution and, when it is not, how the issued command should evolve conservatively.

To address this problem, this paper proposes a \emph{confidence-gated vision-only heading alignment framework} for UAV--UGV systems. The main contributions of this paper are as follows:
\begin{itemize}
    \item A \emph{confidence-gated vision-only heading alignment framework} is proposed for UAV--UGV systems, in which a previously developed heading predictor is treated as a fixed perception module and a lightweight decision layer is introduced on top of it.
    
    \item The decision layer uses two interpretable and computationally lightweight reliability proxies derived directly from the perception stream: a visibility-related proxy based on bounding-box area and a stability-related proxy based on short-window variation in the predicted heading sequence.
    
    \item Heading execution is formulated as an \emph{execute/hold decision problem}, so that predicted headings are issued only when the current perceptual evidence is sufficiently reliable rather than being passed directly to the control system at every step.
    
    \item The study shows that confidence gating should be treated as a \emph{fallback-aware} decision problem: under sparse execution, the baseline freeze-HOLD policy can lead to stale-command accumulation, especially under stronger perturbation.
    
    \item To address this limitation, a \emph{bounded-blend fallback update} is introduced, allowing the issued command to evolve conservatively during low-confidence intervals, reducing stale-command persistence without fully trusting unreliable predictions.
    
    \item The proposed framework is evaluated using both prediction-level and decision-level metrics. In addition to prediction error, the evaluation considers execution rate, issued-command accuracy, command smoothness, and robustness under controlled centroid and area perturbations. This enables the study to distinguish between two related but different questions: whether the model can accurately predict the heading, and whether the system should trust and execute that prediction as a command.
\end{itemize}

The remainder of the paper is organized as follows. Section~\ref{sec:related_work} reviews related work in UAV--UGV cooperation, vision-based landing and following, and uncertainty-aware autonomous decision making. Section~\ref{sec:problem_formulation} formulates the heading prediction and execution problem. Section~\ref{sec:methodology} presents the confidence-gated decision framework and fallback policies. Section~\ref{sec:experiments} describes the experimental setup, and Section~\ref{sec:results} reports quantitative results, robustness analysis, and the comparative evaluation of hold-policy behavior.


\section{Related Work}
\label{sec:related_work}

\subsection{UAV--UGV Cooperative Systems}

Cooperative aerial--ground robotics has emerged as an important research area because UAVs and UGVs provide complementary operational capabilities. UAVs offer rapid viewpoint changes, broad scene coverage, and flexible sensing, whereas UGVs provide persistent ground interaction, payload support, and longer-duration operation \cite{munasinghe2024uavugv, liu2022airground, shahar2025ugvuav, chen2025multirobot}. Recent review literature shows that UAV--UGV collaboration is being studied across applications including environmental monitoring, disaster response, infrastructure inspection, and precision agriculture, while also identifying open challenges in perception, coordination, communication, and uncertainty handling \cite{munasinghe2024uavugv, liu2022airground, ren2024agriculture, shahar2025ugvuav, myndec, myndec2}. These findings support the broader relevance of aerial--ground coordination and motivate reliable relative perception between the two platforms.

Within cooperative UAV--UGV systems, one recurring requirement is the ability of the aerial platform to align with, follow, rendezvous with, or land relative to the ground platform. These tasks depend not only on accurate perception, but also on robust decision making when sensory evidence becomes unreliable. The present work focuses specifically on this perception-to-decision interface and, more narrowly, on how predicted headings should be translated into issued commands under variable confidence.

\subsection{Vision-Based Following and Landing on Moving Ground Platforms}

Vision-based methods are widely used for relative localization and landing because they reduce dependence on external positioning infrastructure and can operate directly from onboard sensing \cite{morales2023vision, falanga2017vision, wu2019autonomous, araar2017landing, feng2018landing, keipour2022visualservoing, cui2023coarse, liu2025uwbvision}. Morales \emph{et al.} present a vision-based method that enables an aerial robot to detect, follow, and land on a moving ground platform, demonstrating the practicality of visual perception for relative motion tasks between aerial and terrestrial robots \cite{morales2023vision}. Falanga \emph{et al.} further show that a quadrotor can autonomously land on a moving platform using only onboard sensing and computing, highlighting the feasibility of high-performance vision-driven landing in dynamic scenarios \cite{falanga2017vision}. Similarly, Wu \emph{et al.} describe an autonomous UAV landing system based on visual navigation, reinforcing the role of image-based geometric information in landing-related control pipelines \cite{wu2019autonomous}. Other studies extend this line of work through marker-based landing, visual servoing, coarse-to-fine visual landing, and integrated sensing frameworks for degraded environments \cite{araar2017landing, feng2018landing, keipour2022visualservoing, cui2023coarse, liu2025uwbvision}.

These studies establish that vision can provide the information required for relative alignment and landing. However, they do not by themselves resolve a separate systems question that arises when visual predictions are fed into a control loop: whether every predicted state or heading estimate should be executed immediately, and how the command stream should evolve when the current estimate is deemed unreliable. In other words, the existence of a visual predictor does not automatically provide a policy for deciding when the current estimate is reliable enough to act upon or how low-confidence intervals should be handled.

\subsection{Uncertainty, Reliability, and Trust in Vision-Based Prediction}

The uncertainty literature in computer vision and machine learning provides strong motivation for distinguishing between predictive output and action reliability. Kendall and Gal show that uncertainty in vision models is not monolithic and instead includes both aleatoric uncertainty, which reflects noise inherent in observations, and epistemic uncertainty, which reflects uncertainty in the model itself \cite{kendall2017uncertainties}. Gal and Ghahramani further show how practical uncertainty estimates can be obtained in deep networks through approximate Bayesian inference with dropout \cite{gal2016dropout}. Guo \emph{et al.} show that modern neural networks can be poorly calibrated even when they achieve strong predictive accuracy, making confidence quality an important issue in downstream decision making \cite{guo2017calibration}. Hendrycks and Gimpel provide an influential baseline for detecting misclassified and out-of-distribution examples, reinforcing the need to reason explicitly about confidence and abnormal inputs \cite{hendrycks2017ood}. More recent work further emphasizes the importance of uncertainty quantification and OOD-aware evaluation in safety-critical perception systems \cite{ovadia2019trust, schwaiger2020uncertainty, araujo2024roadsafety, nuhu2026validation, xue2026unified}.

Lakshminarayanan \emph{et al.} demonstrate that predictive uncertainty estimation is important when neural models are used beyond point prediction, particularly when the system must reason about confidence in its outputs \cite{lakshminarayanan2017deep}. Ovadia \emph{et al.} further show that trustworthiness can degrade substantially under dataset shift even when a model appears accurate under standard test conditions \cite{ovadia2019trust}. This observation is directly relevant to autonomous control settings, where a model may encounter degraded detections or non-nominal visual conditions not fully represented in the original data distribution. Related work on selective prediction and reject-option learning also supports the principle that abstention or deferred action can be preferable when confidence is insufficient \cite{geifman2017selective, geifman2019selectivenet, yildirim2019selective}.

From the perspective of this paper, these works motivate a key design principle: low average prediction error does not imply that every prediction should be executed as a control command. In a heading-alignment context, short-term perception degradation can produce unstable predicted commands even when the underlying predictor remains accurate on average. Moreover, once execution is selectively withheld, the behavior of the issued command stream becomes part of the overall system design, since command quality then depends not only on the gate decision but also on how non-executed intervals are handled \cite{michelmore2018uqdriving, zhang2014failureprediction, araujo2024roadsafety}.

\subsection{Positioning of the Present Work}

A prior work developed the vision-based heading predictor used in the present study \cite{ahmari2025visual}. The present paper is therefore not primarily a new visual predictor, nor does it attempt to estimate full Bayesian uncertainty inside the heading model. Instead, it addresses a distinct and practically important problem: how to introduce a lightweight decision layer between perception and control once heading predictions are already available \cite{zhao2025multimodal, luo2024robotics}.

The proposed framework takes the trained heading predictor as given and adds a confidence-gated decision layer using two interpretable reliability proxies available from the perception stream: a visibility-related proxy derived from bounding-box area and a stability-related proxy derived from short-window variability in predicted heading. In this sense, the work is complementary to prior studies on vision-based following and landing: rather than replacing the perception model, it focuses on the decision interface that determines whether the current predicted heading should be fully executed.

A further contribution of the present work is to treat fallback behavior during low-confidence intervals as part of the decision problem itself. The baseline freeze-HOLD policy provides a simple mechanism for withholding unreliable updates by retaining the previously issued command. However, the analysis shows that this behavior can lead to stale-command accumulation when execution becomes sparse under strong perturbation. To address this limitation, the paper also studies a bounded-blend fallback update that allows the issued command to evolve conservatively during low-confidence intervals without fully trusting the current prediction.

Accordingly, the contribution of the paper lies at the decision level rather than the prediction level. It is conceptually motivated by the broader uncertainty literature, is complementary to prior work on vision-based UAV--UGV interaction, and contributes a lightweight mechanism for deciding not only when a perception-driven heading command should be executed, but also how the issued command should evolve when full execution is withheld.

\section{Problem Formulation}
\label{sec:problem_formulation}

This paper considers a vision-based heading-alignment problem in a UAV--UGV cooperative setting. A previously developed vision-based heading predictor from prior work \cite{ahmari2025visual} is treated as a fixed perception module. This section formulates the vision-based heading-alignment problem considered in this work. It defines the observation and prediction model, the execute/hold decision process, the issued-command update rule, and the problem objective used to characterize decision-level system behavior.

\subsection{Observation and Prediction Model}

Let \(t\) denote the observation index. For each observation, the perception stream provides image-space bounding-box information for the target platform. From this detection, a feature vector
\begin{equation}
\mathbf{x}_t = [cx_t,\; cy_t,\; area_t,\; aspect\_ratio_t]^\top \in \mathbb{R}^{4}
\label{eq:feature_vector}
\end{equation}
is constructed, where \(cx_t\) and \(cy_t\) denote the bounding-box centroid coordinates, \(area_t\) denotes the bounding-box area, and \(aspect\_ratio_t\) denotes the bounding-box aspect ratio.

The ground-truth heading at step \(t\) is denoted by
\[
\theta_t \in \mathbb{R},
\label{eq:ground_truth_heading}
\]
and the frozen heading predictor is represented by
\[
f:\mathbb{R}^{4} \rightarrow \mathbb{R}.
\label{eq:predictor_mapping}
\]
The predicted heading is therefore
\[
\hat{\theta}_t = f(\mathbf{x}_t).
\label{eq:predicted_heading}
\]

In this work, \(f(\cdot)\) is treated as a fixed, pre-trained heading predictor whose architecture, training, and standalone performance are reported in prior work \cite{ahmari2025visual}. The contribution of this paper lies in the decision layer that operates on top of the predictor outputs.

\subsection{Execute/Hold Decision Process}

At each step \(t\), the system must decide whether the current predicted heading should be issued directly as a control command. This decision is represented by the action variable
\begin{equation}
a_t \in \{\mathrm{EXECUTE},\; \mathrm{HOLD}\}.
\label{eq:action_variable}
\end{equation}

If \(a_t = \mathrm{EXECUTE}\), the current predicted heading is issued directly to the control system. If \(a_t = \mathrm{HOLD}\), the system does not fully commit to the current prediction; Instead, \textbf{this work introduces a conservative fallback update rule} that prevents indefinite command freezing by allowing the issued command to evolve cautiously during prolonged low-confidence intervals.

Accordingly, this paper distinguishes between the predictor output sequence
\[
\{\hat{\theta}_t\}_{t=1}^{T}
\label{eq:prediction_sequence}
\]
and the issued command sequence
\[
\{u_t\}_{t=1}^{T},
\label{eq:issued_sequence}
\]
where \(u_t\) denotes the command actually passed to the control system.

\subsection{Issued-Command Update}

The issued command \(u_t\) depends not only on the current prediction \(\hat{\theta}_t\), but also on the action \(a_t\) and the recent hold history. Let \(h_t\) denote the number of consecutive HOLD steps up to time \(t\). The issued-command update is defined as
\begin{equation}
u_t =
\begin{cases}
\hat{\theta}_t, &  a_t = \mathrm{EXECUTE},\\[4pt]
(1-\alpha)u_{t-1} + \alpha \hat{\theta}_t, &  a_t = \mathrm{HOLD}\ \text{and}\ h_t \le \bar{h},\\[4pt]
(1-\rho)u_{t-1} + \rho \hat{\theta}_t, &  a_t = \mathrm{HOLD}\ \text{and}\ h_t > \bar{h},
\end{cases}
\label{eq:issued_command_update}
\end{equation}
where \(\alpha\) is a short-hold blending factor, \(\bar{h}\) is the maximum bounded-hold duration, and \(\rho\) is a recovery blending factor for prolonged low-confidence intervals.

Equation~\eqref{eq:issued_command_update} reflects the intended behavior of the system. High-confidence observations are executed directly, whereas low-confidence observations do not trigger full execution but still allow the issued command to evolve conservatively. For reference, the baseline freeze-HOLD can be written as
\begin{equation}
u_t = u_{t-1}, \qquad \text{if } a_t = \mathrm{HOLD},
\label{eq:freeze_hold_baseline}
\end{equation}
which indefinitely repeats the previous command during non-execution intervals. In contrast, the bounded-blend rule in \eqref{eq:issued_command_update} mitigates stale-command persistence when the gate remains closed for multiple consecutive steps.

This formulation explicitly separates three components of the overall system:
\begin{itemize}
    \item The \emph{perception problem}, handled by the fixed heading predictor \(f(\cdot)\).
    \item The \emph{decision problem}, which determines whether the current prediction should be fully executed, and
    \item the \emph{fallback command-update problem}, which determines how the issued command evolves during low-confidence intervals.
\end{itemize}

\subsection{Problem Objective}

The objective of the proposed framework is to design a decision mechanism that determines when the output of the existing heading predictor should be executed and how the issued command should evolve when full execution is not justified. The desired behavior is to balance three requirements:
\begin{enumerate}
    \item preserve the usefulness of the inherited heading predictor,
    \item execute only sufficiently reliable predicted commands, and
    \item maintain accurate and smooth issued-command behavior under nominal and perturbed perception conditions.
\end{enumerate}

Accordingly, the overall problem addressed in this paper is not only one of heading prediction, but one of \emph{prediction-conditioned action selection}: given a predicted heading \(\hat{\theta}_t\), determine whether it should be executed directly or incorporated through a conservative fallback update during low-confidence intervals.

\section{Methodology}
\label{sec:methodology}

This section presents the proposed decision framework that operates on top of the frozen heading predictor and transforms its output into a more selective and reliability-aware command-issuance process. The methodology includes interpretable reliability proxies derived from the perception stream, a confidence-gated execute/hold policy, a comparison between the baseline freeze-HOLD policy and the bounded-blend HOLD policy during low-confidence intervals, a calibration-based threshold-selection procedure, and an evaluation protocol for decision-level robustness and smoothness.




\subsection{Overview of the Decision Layer}

The proposed framework takes as input the predicted heading sequence generated by the frozen perception module and evaluates whether the current estimate is sufficiently reliable for direct execution. The decision process uses two complementary reliability proxies:
\begin{itemize}
    \item A \emph{visibility-related proxy} based on the detected target's bounding-box area in image coordinates, and
    \item a \emph{stability-related proxy} based on short-window variation in predicted headings.
\end{itemize}

The first proxy reflects how strongly the target is represented in the image through its detected bounding-box size, while the second reflects the short-term consistency of the predicted heading sequence. Heading commands are executed only when the target is sufficiently visible and the recent predictions are sufficiently stable.

In this paper, a \emph{low-confidence interval} refers to one or more consecutive time steps at which the reliability conditions for direct execution are not satisfied. Concretely, such intervals occur when the visibility proxy falls below its threshold and/or the stability proxy exceeds its threshold, so the gate selects \(\mathrm{HOLD}\) rather than \(\mathrm{EXECUTE}\), as formalized later in \eqref{eq:gate_policy}.

If either condition is violated, the system assigns the action \(a_t=\mathrm{HOLD}\) rather than fully executing the current prediction. In that case, the issued command does not remain undefined; instead, it is updated according to the fallback rule in \eqref{eq:issued_command_update}, which determines \(u_t\) from the previous issued command, the current predicted heading, and the recent hold history. In this way, the framework separates two roles: the reliability proxies determine whether the current prediction should be executed, while the fallback update determines how the issued command evolves during low-confidence intervals.

\subsection{Visibility-Related Reliability Proxy}

The first reliability proxy is based on the area of the detected target bounding box in image coordinates.

\begin{figure}[!h]
    \centering
    \includegraphics[width=\columnwidth]{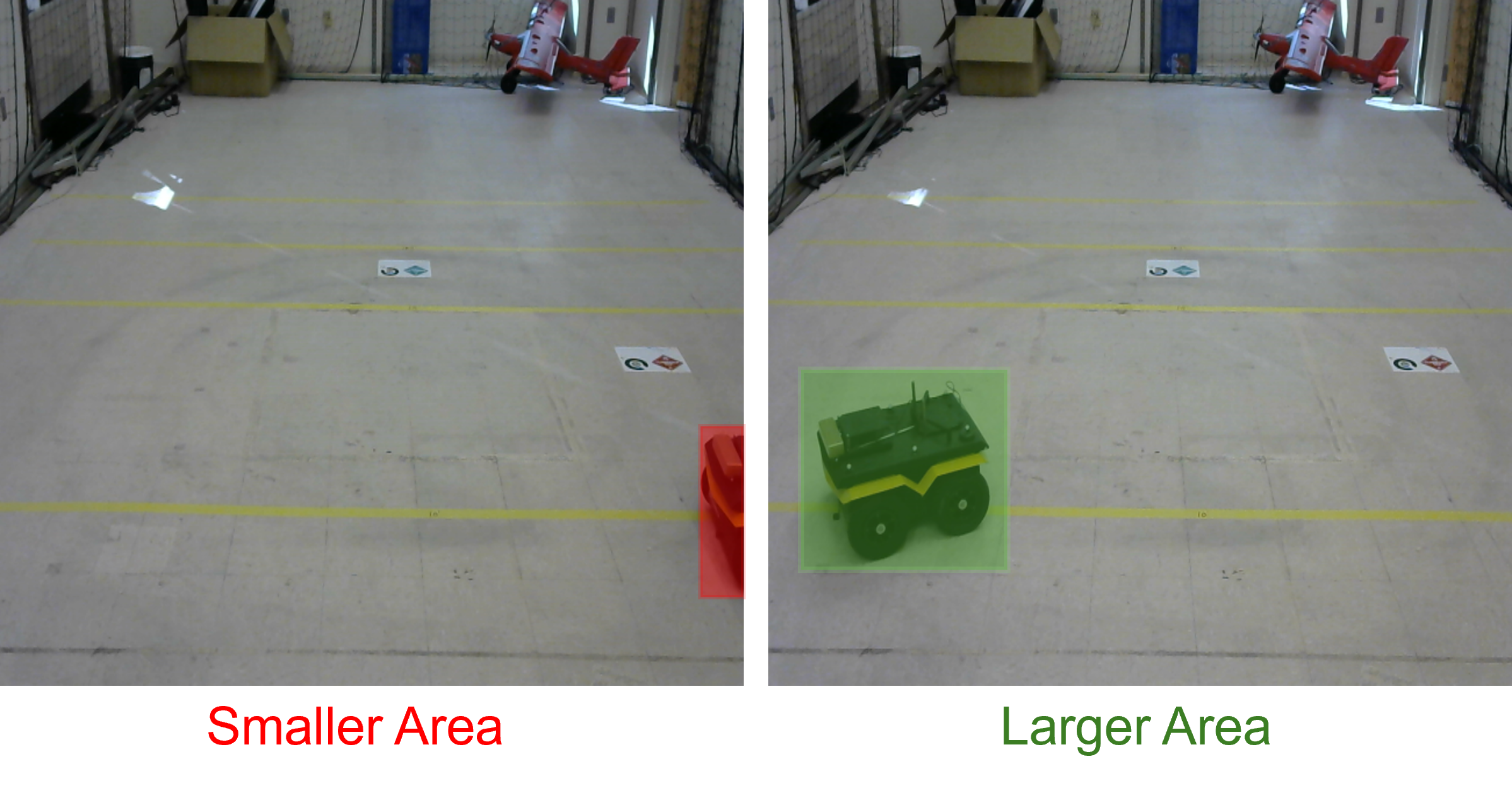}
    \caption{Illustration of the visibility-related proxy based on bounding-box area. A \textit{smaller} detected area corresponds to weaker apparent target scale in the image, whereas a \textit{larger} detected area indicates stronger visual presence.}
    \label{fig:area_visibility_proxy}
\end{figure}

In the proposed framework, \(area_t\) is used as a visibility-related proxy. A smaller area generally corresponds to weaker apparent target scale in the image and therefore potentially less reliable visual evidence, whereas a larger area indicates stronger visual presence, as illustrated in Fig.~\ref{fig:area_visibility_proxy}. The role of this quantity in the decision layer is not to serve as a geometric state estimate, but to provide a simple measure of whether the current observation is likely to be informative enough for safe command execution.

\subsection{Stability-Related Reliability Proxy}

The second reliability proxy is based on short-window consistency in the predicted heading sequence. For a window size \(k \ge 2\), define the rolling mean
\begin{equation}
\bar{\theta}_t = \frac{1}{k}\sum_{j=0}^{k-1}\hat{\theta}_{t-j},
\label{eq:rolling_mean}
\end{equation}
and the rolling population standard deviation
\begin{equation}
\sigma_{\theta}(t) =
\sqrt{
\frac{1}{k}\sum_{j=0}^{k-1}
\left(\hat{\theta}_{t-j} - \bar{\theta}_t\right)^2
},
\qquad t \ge k.
\label{eq:rolling_std}
\end{equation}

The quantity \(\sigma_{\theta}(t)\) measures local variability in the recent predicted-heading sequence. A small value indicates that recent predictions are mutually consistent, whereas a larger value indicates short-term fluctuation. In this work, \(\sigma_{\theta}(t)\) is interpreted as a stability-related reliability signal for decision making.

\subsection{Confidence-Gated Execute/Hold Policy}

Using the two reliability proxies, the decision layer applies a confidence-gated rule with thresholds \(\tau_{area}\) and \(\tau_{\sigma}\):
\begin{equation}
a_t =
\begin{cases}
\mathrm{EXECUTE}, & \text{if } area_t \ge \tau_{area}
\;\wedge\;
\sigma_{\theta}(t) \le \tau_{\sigma},\\[4pt]
\mathrm{HOLD}, & \text{otherwise.}
\end{cases}
\label{eq:gate_policy}
\end{equation}

Equation~\eqref{eq:gate_policy} reflects the intended behavior of the system:
\begin{itemize}
    \item \textbf{Execute} only when the target is sufficiently visible and the predicted heading is sufficiently stable;
    \item \textbf{Hold} when visual evidence is weak or when the recent predicted-heading sequence is unstable.
\end{itemize}

The resulting issued command sequence follows the update rule defined in \eqref{eq:issued_command_update}: EXECUTE issues the current predicted heading directly, whereas HOLD triggers a bounded fallback update rather than indefinite freezing.

\subsection{Fallback Behavior During HOLD}

A HOLD decision indicates that the current perceptual evidence is not reliable enough for full command execution. In the baseline freeze-HOLD formulation, HOLD caused the system to indefinitely retain the previously issued command, as described by \eqref{eq:freeze_hold_baseline}. While this strategy provides a simple baseline for suppressing unreliable updates, it can also produce stale-command accumulation when the gate remains closed for multiple consecutive steps under strong perturbation.

To address this issue, the present framework adopts the bounded-blend fallback defined in \eqref{eq:issued_command_update}. This design allows the issued command to evolve conservatively during low-confidence intervals without fully trusting the current prediction. As a result, the framework preserves decision-level selectivity while mitigating the stale-command persistence that can occur during prolonged baseline freeze-HOLD behavior.

\subsection{Threshold Selection}

The confidence gate depends on two thresholds: the visibility threshold \(\tau_{area}\) and the stability threshold \(\tau_{\sigma}\). These thresholds are selected on a calibration subset rather than the final test subset in order to avoid direct tuning on evaluation data.

In the present study, the thresholds are selected using a Pareto-grid search over candidate values of \(\tau_{area}\) and \(\tau_{\sigma}\) \cite{deb2011multi, miettinen1999nonlinear,emmerich2018tutorial}. Specifically, a predefined grid of threshold pairs \((\tau_{area}, \tau_{\sigma})\) is evaluated on the calibration subset, and each candidate operating point is assessed in terms of:
\begin{itemize}
    \item The fraction of executed observations,
    \item the prediction accuracy on executed observations, and
    \item the smoothness of the resulting issued command stream.
\end{itemize}

The final operating point is selected from this grid as a data-driven trade-off among these competing objectives. Once selected, the same threshold pair is fixed for all subsequent clean and perturbed evaluations reported in the paper.

\subsection{Comparative Gate Variants}

To isolate the role of each reliability proxy, four execution strategies are considered:
\begin{enumerate}
    \item \textbf{Always-execute baseline:} every predicted heading is issued.
    \item \textbf{Area-only gate:} a heading is executed only if \(area_t \ge \tau_{area}\).
    \item \textbf{Stability-only gate:} a heading is executed only if \(\sigma_{\theta}(t) \le \tau_{\sigma}\).
    \item \textbf{Full confidence gate:} a heading is executed only if both conditions are satisfied.
\end{enumerate}

For the gated strategies, the paper evaluates two fallback behaviors during HOLD:
\begin{itemize}
    \item The \emph{baseline freeze-HOLD policy} \eqref{eq:freeze_hold_baseline}, and
    \item the \emph{bounded-blend HOLD policy} \eqref{eq:issued_command_update}.
\end{itemize}

This comparison makes it possible to determine not only whether the observed effects arise primarily from the visibility proxy, the stability proxy, or their combination, but also how strongly the final system behavior depends on the fallback command-update design once execution becomes sparse.

\subsection{Evaluation Metrics}

The proposed framework is intended for decision-level improvement rather than predictor redesign. Accordingly, evaluation is performed both at the predictor level and at the decision level. Predictor-level performance is measured using standard regression metrics applied to the inherited heading estimates \(\hat{\theta}_t\), while decision-level performance is measured using metrics derived from the execute/hold actions and the issued command sequence \(u_t\).

Let \(T\) denote the number of evaluated observations. The mean absolute error is defined as
\begin{equation}
\mathrm{MAE} = \frac{1}{T}\sum_{t=1}^{T}\left|\theta_t - \hat{\theta}_t\right|,
\label{eq:mae}
\end{equation}
and the root mean squared error is defined as
\begin{equation}
\mathrm{RMSE} =
\sqrt{
\frac{1}{T}\sum_{t=1}^{T}\left(\theta_t - \hat{\theta}_t\right)^2
}.
\label{eq:rmse}
\end{equation}

However, predictor-level error alone is not sufficient to characterize the behavior of the overall system, because the final control behavior depends on the issued command sequence \(u_t\), not only on the raw predictions \(\hat{\theta}_t\). For this reason, the framework is also evaluated using the decision-level metrics defined below.

\paragraph{Executed fraction}
Let \(\mathbb{I}[\cdot]\) denote the indicator function. The executed fraction is defined as
\begin{equation}
\mathrm{ExecFrac} = \frac{1}{T}\sum_{t=1}^{T}\mathbb{I}[a_t = \mathrm{EXECUTE}],
\label{eq:executed_fraction}
\end{equation}
which measures the proportion of observations for which the current predicted heading is issued directly.

\paragraph{MAE on executed observations only}
Let
\[
\mathcal{E} = \{\, t \in \{1,\dots,T\} : a_t = \mathrm{EXECUTE} \,\}
\]
denote the set of executed observations, and let \(|\mathcal{E}|\) denote its cardinality. The mean absolute error on executed observations is defined as
\begin{equation}
\mathrm{MAE}_{\mathrm{exec}} =
\frac{1}{|\mathcal{E}|}
\sum_{t \in \mathcal{E}}
\left|\theta_t - \hat{\theta}_t\right|.
\label{eq:mae_executed_only}
\end{equation}
This metric evaluates the quality of the subset of predictions that the gate allows to be fully executed.

\paragraph{MAE of the issued command stream}
Because the final control behavior depends on the issued command sequence \(u_t\), the issued-stream mean absolute error is defined as
\begin{equation}
\mathrm{MAE}_{\mathrm{issued}} =
\frac{1}{T}\sum_{t=1}^{T}\left|\theta_t - u_t\right|.
\label{eq:mae_issued_stream}
\end{equation}
This metric reflects the accuracy of the command actually passed to the control system rather than the raw predictor output alone.

\paragraph{Mean jerk}
To quantify smoothness of the issued command stream, the second-difference magnitude is defined as
\begin{equation}
J(t) = |u_t - 2u_{t-1} + u_{t-2}|, \qquad t \ge 3.
\label{eq:jerk_metric}
\end{equation}
The reported smoothness statistic is the mean jerk,
\begin{equation}
\overline{J} = \frac{1}{T-2}\sum_{t=3}^{T} J(t),
\label{eq:mean_jerk}
\end{equation}
where lower values indicate smoother evolution of the issued command stream.

Together, these metrics characterize the main decision-level trade-off studied in this paper: predictor accuracy, execution availability, executed-subset quality, issued-command accuracy, and command smoothness.

\subsection{Robustness Evaluation Under Perception Perturbation}

To study the behavior of the decision layer under degraded perception, controlled perturbations are applied to the input features during evaluation, while ground-truth headings remain unchanged. Two perturbation modes are considered.

\paragraph{Centroid perturbation}
The centroid coordinates are perturbed additively:
\begin{equation}
cx'_t = cx_t + \epsilon_t^{(cx)}, 
\qquad
cy'_t = cy_t + \epsilon_t^{(cy)},
\label{eq:centroid_perturbation}
\end{equation}
with
\[
\epsilon_t^{(cx)},\; \epsilon_t^{(cy)} \sim \mathcal{N}(0,\sigma_c^2).
\label{eq:centroid_noise}
\]

\paragraph{Area perturbation}
The area feature is perturbed multiplicatively:
\begin{equation}
area'_t = area_t(1 + \epsilon_t^{(a)}),
\label{eq:area_perturbation}
\end{equation}
with
\[
\epsilon_t^{(a)} \sim \mathcal{N}(0,\sigma_a^2).
\label{eq:area_noise}
\]

These perturbations simulate perception variability without modifying the inherited heading predictor or the ground-truth labels. The resulting experiments evaluate how the execute/hold policy changes as uncertainty increases and how that affects executed-frame accuracy, issued-command accuracy, and command smoothness.


\section{Experimental Setup}
\label{sec:experiments}
The experiments evaluate the decision framework defined in Sections~\ref{sec:problem_formulation} and \ref{sec:methodology} on a UAV--UGV dataset with ground-truth heading labels generated using a VICON motion-capture system. The inherited heading predictor is kept fixed throughout all experiments so that observed performance differences are attributable to the decision layer and fallback policy rather than to changes in the perception model.

 The experiments are designed to answer three questions:
\begin{enumerate}
    \item How does the confidence-gated framework behave on the clean sequence relative to always-execute?
    \item How do the gate variants respond to centroid and area perturbations under the baseline freeze-HOLD policy?
    \item To what extent does the bounded-blend fallback mitigate stale-command behavior once execution becomes sparse?
\end{enumerate}

\subsection{Data Partition and Calibration Protocol}

The dataset used in this study is the real UAV--UGV visual-heading dataset originally collected and described in \cite{ahmari2025visual}. The data were collected in an indoor laboratory environment using a VICON motion-capture system to provide millimeter-level position ground truth. To avoid tuning the decision layer on final evaluation data, the available ordered sequence is divided into a calibration subset and a test subset. The calibration subset is used only to select the gate thresholds, and \textit{all final results} are reported on the test subset.

Threshold calibration follows the Pareto-style grid-search procedure described in Section~\ref{sec:methodology}. Candidate threshold pairs \((\tau_{area}, \tau_{\sigma})\) are evaluated on the calibration subset, and the selected operating point is then fixed for all subsequent clean and perturbed evaluations. No threshold re-tuning is performed on the test sequence or at individual perturbation levels.

\subsection{Evaluated Execution Strategies}

Four gate-level execution strategies are evaluated:
\begin{enumerate}
    \item \textbf{Always-execute baseline}
    \item \textbf{Area-only gate}
    \item \textbf{Stability-only gate}
    \item \textbf{Full confidence gate}
\end{enumerate}

For the gated strategies, two HOLD behaviors are considered:
\begin{itemize}
    \item The baseline freeze-HOLD policy, and
    \item The bounded-blend HOLD policy.
\end{itemize}

This comparison makes it possible to distinguish between the effect of execution selectivity and the effect of fallback behavior once execution becomes sparse.

\subsection{Evaluation Protocol}

All methods are evaluated under clean conditions and under the two perturbation modes defined in Section~\ref{sec:methodology}: centroid perturbation and area perturbation. For each condition, the predictor output, gate decision, and issued-command stream are recomputed using the same fixed predictor and the same fixed threshold pair.

The reported quantities include predictor-level accuracy together with the decision-level metrics defined in Section~\ref{sec:methodology}, namely executed fraction, executed-only accuracy, issued-command accuracy, and issued-command smoothness.

For the perturbation experiments, repeated Monte Carlo trials are used to reduce sensitivity to a single random realization and to summarize robustness trends across noise levels.


\section{Results and Discussion}
\label{sec:results}

This section evaluates the proposed confidence-gated decision framework under clean and perturbed perception conditions. This section is organized in two stages. First, the baseline freeze-HOLD policy is analyzed to characterize the behavior of the baseline confidence-gated system and to identify its limitations under sparse execution. Second, the updated bounded-blend HOLD policy is evaluated under the same predictor, threshold pair, and perturbation settings in order to determine whether it mitigates stale-command accumulation without changing the underlying gate decisions. The discussion therefore distinguishes between the effect of \emph{execution selectivity} and the effect of the \emph{fallback command-update rule} once execution becomes sparse.

\subsection{Threshold Calibration}

The threshold pair used throughout all experiments was obtained from the calibration subset using the Pareto grid-search procedure described in Section~\ref{sec:methodology}. For each candidate threshold pair \((\tau_{area},\tau_{\sigma})\) in the predefined search grid, the confidence gate in \eqref{eq:gate_policy} was applied on the calibration sequence, and the resulting operating point was evaluated in terms of executed fraction, executed-only MAE, and issued-command smoothness. This produced a set of candidate trade-off points between execution availability and command quality.

The final threshold pair was selected from this candidate set as a calibration-based operating point that preserved high execution availability while improving the quality of the executed subset and maintaining acceptable smoothness of the issued command stream. Here, \(\tau_{area}\) is dimensionless because \(area_t\) is computed as the bounding-box area normalized by the total image area, while \(\tau_{\sigma}\) is expressed in degrees because \(\sigma_{\theta}(t)\) is computed from predicted heading angles. The selected operating point corresponds to
\begin{equation}
\tau_{area}=0.025583 \;, 
\qquad 
\tau_{\sigma}=0.256323^\circ.
\label{eq:selected_thresholds}
\end{equation}

Once selected on the calibration subset, this threshold pair was fixed for all subsequent clean and perturbed evaluations in both the freeze-HOLD and bounded-blend studies. Importantly, the calibration outputs are identical across the two HOLD policies in terms of selected threshold pair, executed fraction, and executed-only accuracy. This confirms that the comparison between the baseline freeze-HOLD and updated HOLD behaviors isolates the effect of the fallback command-update rule rather than changes in the predictor, threshold selection, or gate decision logic.

\subsection{Baseline Freeze-HOLD Policy}

\subsubsection{Clean-Sequence Behavior}

\begin{table*}[h] \caption{Clean-sequence comparison under the baseline freeze-HOLD policy} \label{tab:clean_results_freeze} \centering \begin{tabular}{lcccc} \hline Mode & Execution Rate (\%) & MAE Executed (deg) & MAE Issued (deg) & Mean Jerk (deg/step$^2$) \\ \hline Always-execute & 100.00 & 0.094285 & 0.094285 & 0.018487 \\ Area-only & 98.78 & 0.091493 & 0.092446 & 0.017173 \\ Stability-only & 97.24 & 0.089771 & 0.102562 & 0.026114 \\ Full gate & 96.35 & 0.088549 & 0.102184 & 0.025597 \\ \hline \end{tabular} \end{table*}

Table~\ref{tab:clean_results_freeze} summarizes the clean-sequence performance of the baseline freeze-HOLD policy. Under nominal conditions, the always-execute baseline issues commands at every step, whereas the gated strategies execute slightly fewer observations: 0.987844 for area-only, 0.972447 for stability-only, and 0.963533 for the full gate. Thus, even on the clean sequence, the confidence gate behaves selectively rather than acting as a degenerate always-execute rule.

At the same time, executed-only MAE improves as the gate becomes more selective. Relative to always-execute, area-only reduces executed-only MAE from 0.094285 to 0.091493, stability-only reduces it further to 0.089771, and the full gate achieves the lowest executed-only MAE of 0.088549. This indicates that even under clean conditions, the gate excludes a subset of observations that are less favorable for direct command execution.

However, the issued-stream behavior of the baseline freeze-HOLD policy is mixed. Area-only slightly improves the baseline issued-stream MAE from 0.094285 to 0.092446 and also reduces mean jerk from 0.018487 to 0.017173. In contrast, the more selective stability-only and full gates increase issued-stream MAE to 0.102562 and 0.102184, respectively, while also increasing mean jerk to 0.026114 and 0.025597. Therefore, under nominal conditions, mild gating produces the best overall balance, whereas stronger withholding does not improve the final issued command stream when HOLD is implemented as indefinite freezing.

\subsubsection{Response to Centroid Perturbation}

\begin{figure}[h]
    \centering
    \includegraphics[width=\columnwidth]{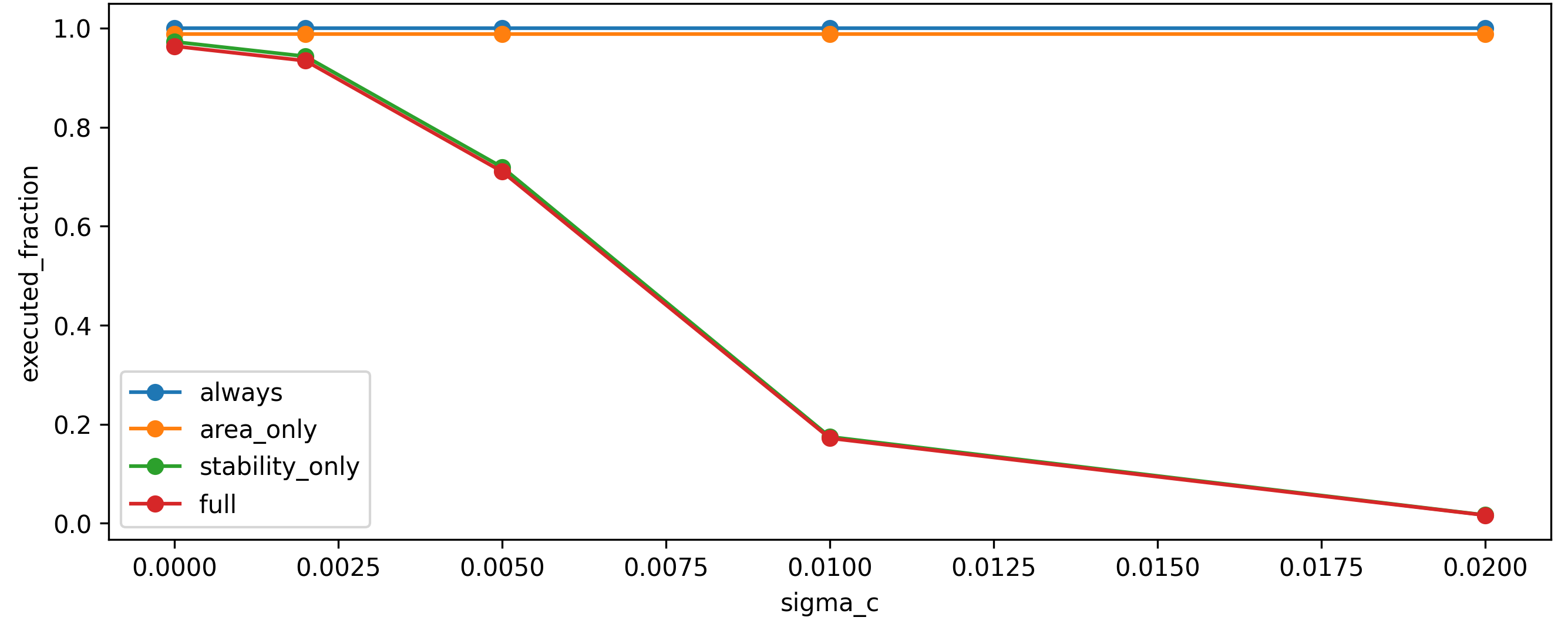}
    \caption{Baseline freeze-HOLD policy: executed fraction versus centroid perturbation magnitude. Stability-only and full gating become highly selective as centroid noise increases, whereas always-execute and area-only remain nearly unchanged.}
    \label{fig:orig_centroid_exec_frac}
\end{figure}

Centroid perturbation exposes the main limitation of the baseline freeze-HOLD policy. Figure~\ref{fig:orig_centroid_exec_frac} shows that always-execute remains fixed at full execution and area-only remains nearly unchanged across the full perturbation range, whereas stability-only and full gating become highly selective as centroid noise increases. Their executed fraction decreases only mildly at very small perturbation, drops sharply by \(\sigma_c=0.005\), falls to approximately 0.17 by \(\sigma_c=0.01\), and collapses to approximately 0.016 by \(\sigma_c=0.02\). This confirms that centroid perturbation primarily affects the stability-related reliability proxy rather than the area proxy.

\begin{figure}[h]
    \centering
    \includegraphics[width=\columnwidth]{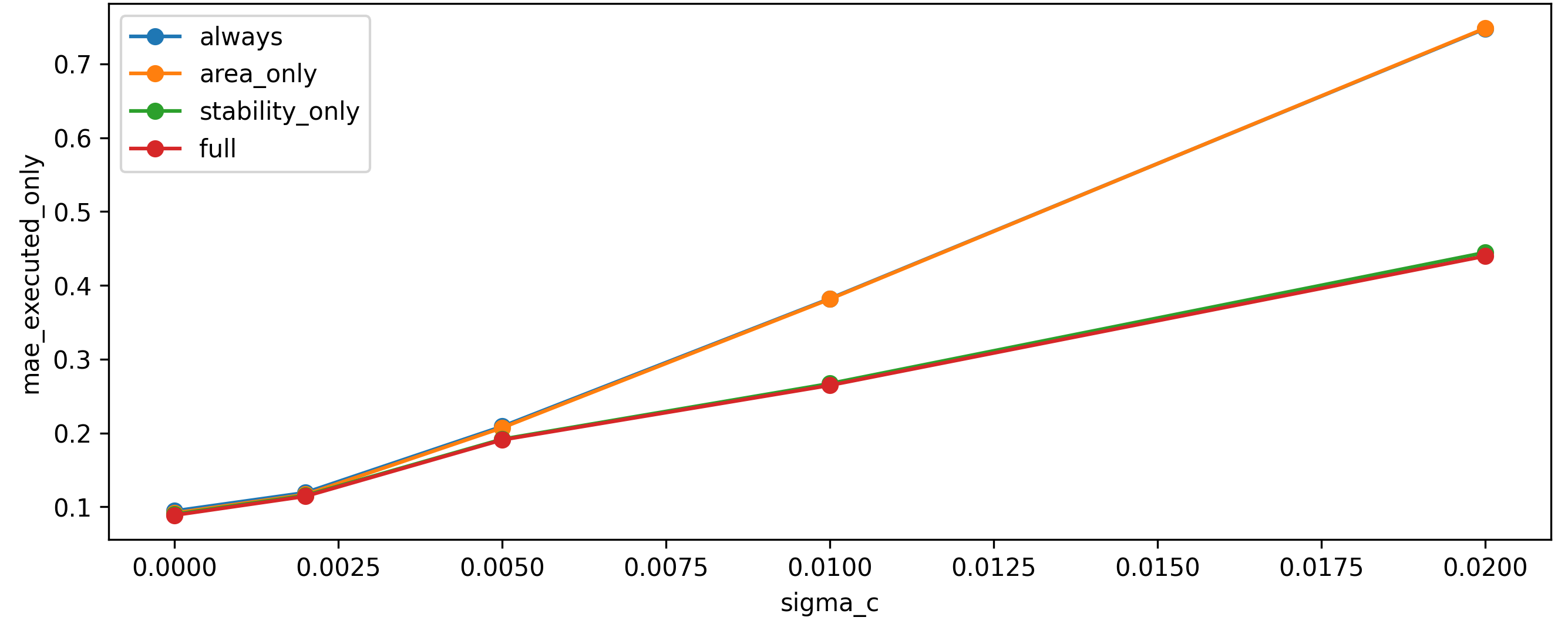}
    \caption{Baseline freeze-HOLD policy: executed-only MAE versus centroid perturbation magnitude. Stability-only and full gating preserve a cleaner executed subset at high centroid noise.}
    \label{fig:orig_centroid_mae_exec}
\end{figure}

Figure~\ref{fig:orig_centroid_mae_exec} reports MAE on the executed subset. At the highest centroid perturbation level, always-execute and area-only both degrade to approximately 0.748, whereas stability-only and the full gate remain near 0.444 and 0.440, respectively. Thus, when the stability-based gates do execute under strong centroid noise, they preserve a substantially cleaner executed subset.

\begin{figure}[!h]
    \centering
    \includegraphics[width=\columnwidth]{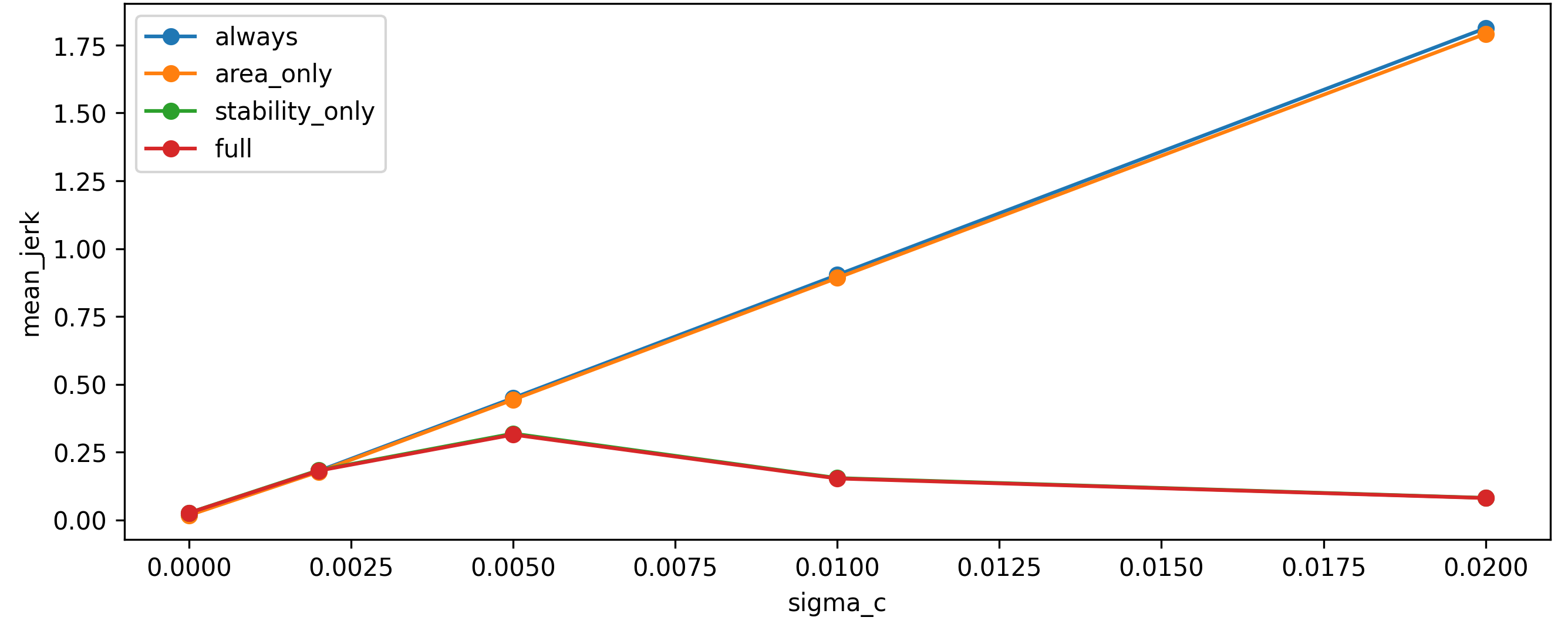}
    \caption{Baseline freeze-HOLD policy: mean jerk versus centroid perturbation magnitude. Stability-only and full gating remain very smooth at high centroid noise, but this smoothness is associated with sparse execution and stale-command persistence.}
    \label{fig:orig_centroid_jerk}
\end{figure}

Figure~\ref{fig:orig_centroid_jerk} shows the corresponding smoothness behavior. For always-execute and area-only, mean jerk rises rapidly with centroid noise and exceeds 1.79 at \(\sigma_c=0.02\). In contrast, stability-only and the full gate reduce mean jerk to approximately 0.08 at the same perturbation level. On its own, this would suggest that strong gating produces a smoother command stream. 

\begin{figure}[h]
    \centering
    \includegraphics[width=\columnwidth]{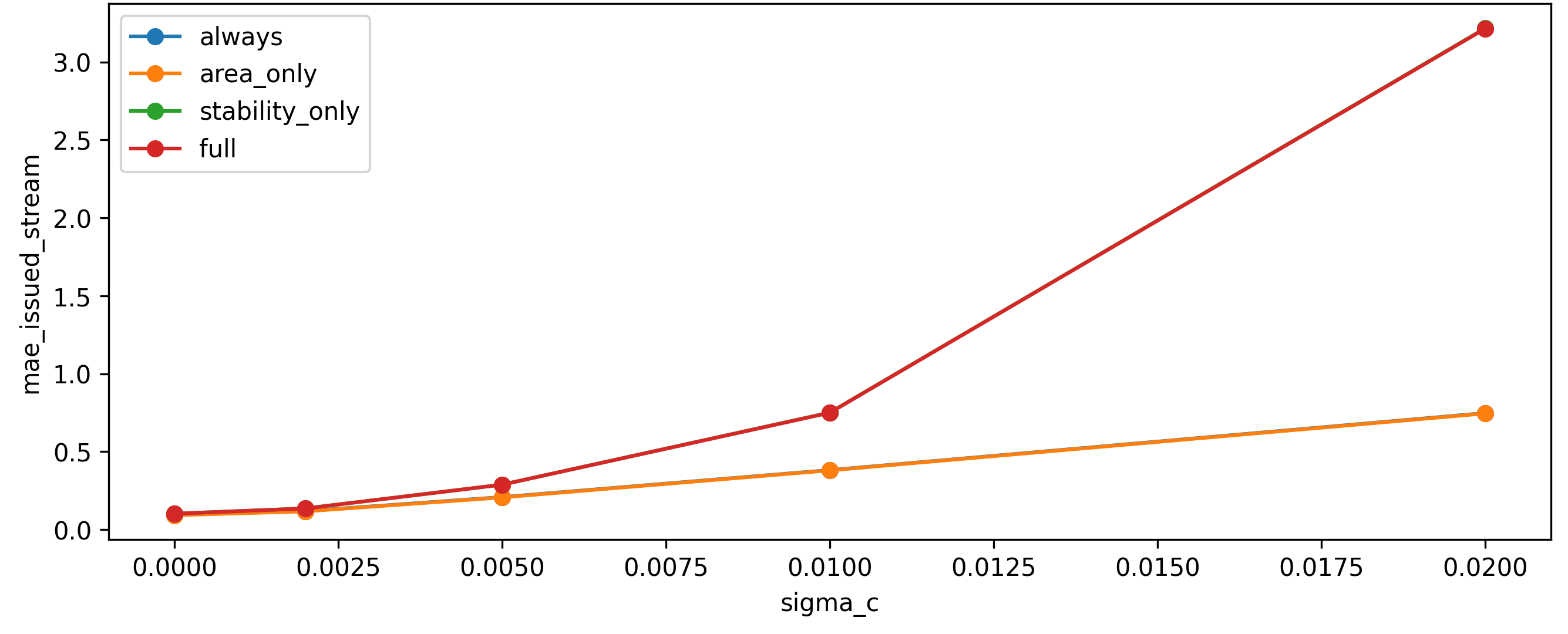}
    \caption{Baseline freeze-HOLD policy: issued-stream MAE versus centroid perturbation magnitude. Stability-only and full gating exhibit severe issued-stream degradation once execution becomes sparse.}
    \label{fig:orig_centroid_mae_issued}
\end{figure}

However, Figure~\ref{fig:orig_centroid_mae_issued} reveals the critical failure mode: issued-stream MAE becomes extremely large once execution becomes too sparse.

Specifically, at \(\sigma_c=0.02\), the freeze-HOLD stability-only and full gates produce issued-stream MAE values of 3.2176 and 3.2152, respectively, far worse than always-execute and area-only. This behavior is summarized in Table~\ref{tab:centroid_high_noise_freeze}. The combination of very low jerk and very large issued-stream error indicates that the issued command becomes stale rather than simply smooth. In other words, the gate successfully suppresses unreliable executions, but freeze-HOLD causes the command stream to remain fixed for too long once updates become rare.

\begin{table}[h] \caption{Baseline freeze-HOLD policy at highest centroid perturbation level \((\sigma_c = 0.02)\)} \label{tab:centroid_high_noise_freeze} \centering \setlength{\tabcolsep}{3.0pt} \small \begin{tabular}{lcccc} \toprule Mode & \makecell{Execution\\Rate (\%)} & \makecell{MAE Exec.\\Only (deg)} & \makecell{MAE\\Issued (deg)} & \makecell{Mean Jerk\\(deg/step$^2$)} \\ \midrule Always-execute & 100.00 & 0.7477 & 0.7477 & 1.8128 \\ Area-only & 98.78 & 0.7484 & 0.7457 & 1.7921 \\ Stability-only & 1.66 & 0.4445 & 3.2177 & 0.0812 \\ Full gate & 1.63 & 0.4397 & 3.2153 & 0.0805 \\ \bottomrule \end{tabular} \end{table}
\subsubsection{Response to Area Perturbation}

\begin{figure}[!h]
    \centering
    \includegraphics[width=0.95\columnwidth]{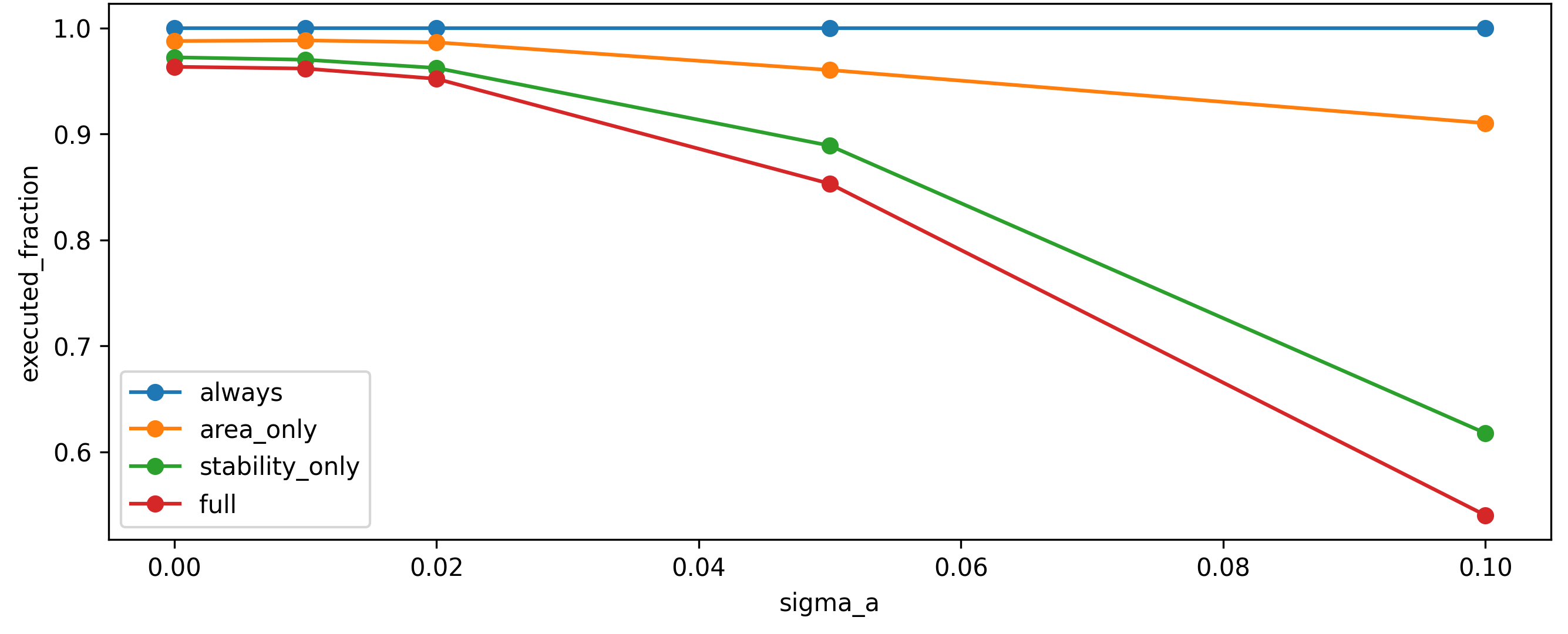}
    \caption{Baseline freeze-HOLD policy: executed fraction versus area perturbation magnitude. The full gate becomes the most selective as area noise increases.}
    \label{fig:orig_area_exec_frac}
\end{figure}

The response to area perturbation is qualitatively different. Figure~\ref{fig:orig_area_exec_frac} shows that always-execute remains at full availability, while the gated strategies become progressively more selective as \(\sigma_a\) increases. Area-only remains relatively permissive, decreasing only to about 0.91 at \(\sigma_a=0.10\). Stability-only decreases to approximately 0.618, and the full gate becomes the most selective, dropping to approximately 0.540. This behavior is consistent with the role of area in the gate design, since the perturbation acts directly on the visibility-related reliability proxy.

\begin{figure}[h]
    \centering
    \includegraphics[width=0.95\columnwidth]{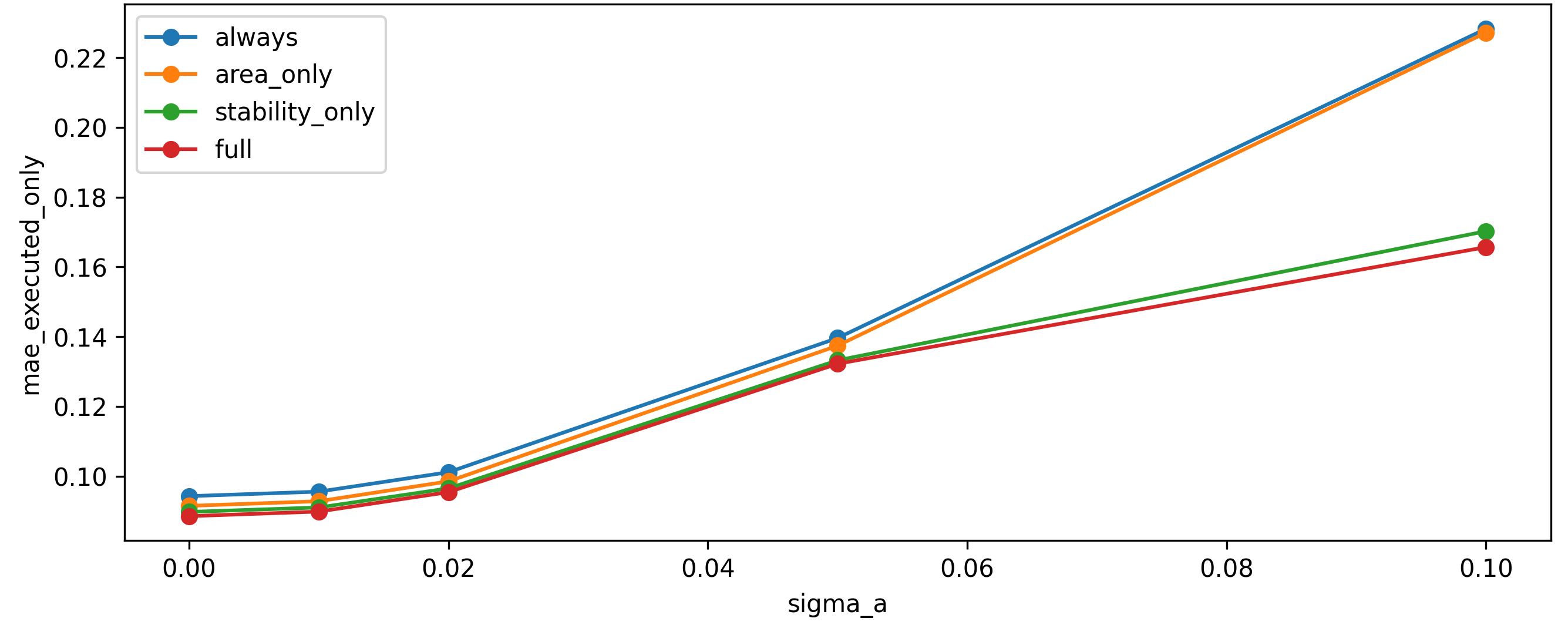}
    \caption{Baseline freeze-HOLD policy: executed-only MAE versus area perturbation magnitude. Stability-only and full gating preserve a cleaner executed subset at high area noise.}
    \label{fig:orig_area_mae_exec}
\end{figure}

\begin{figure}[h]
    \centering
    \includegraphics[width=0.95\columnwidth]{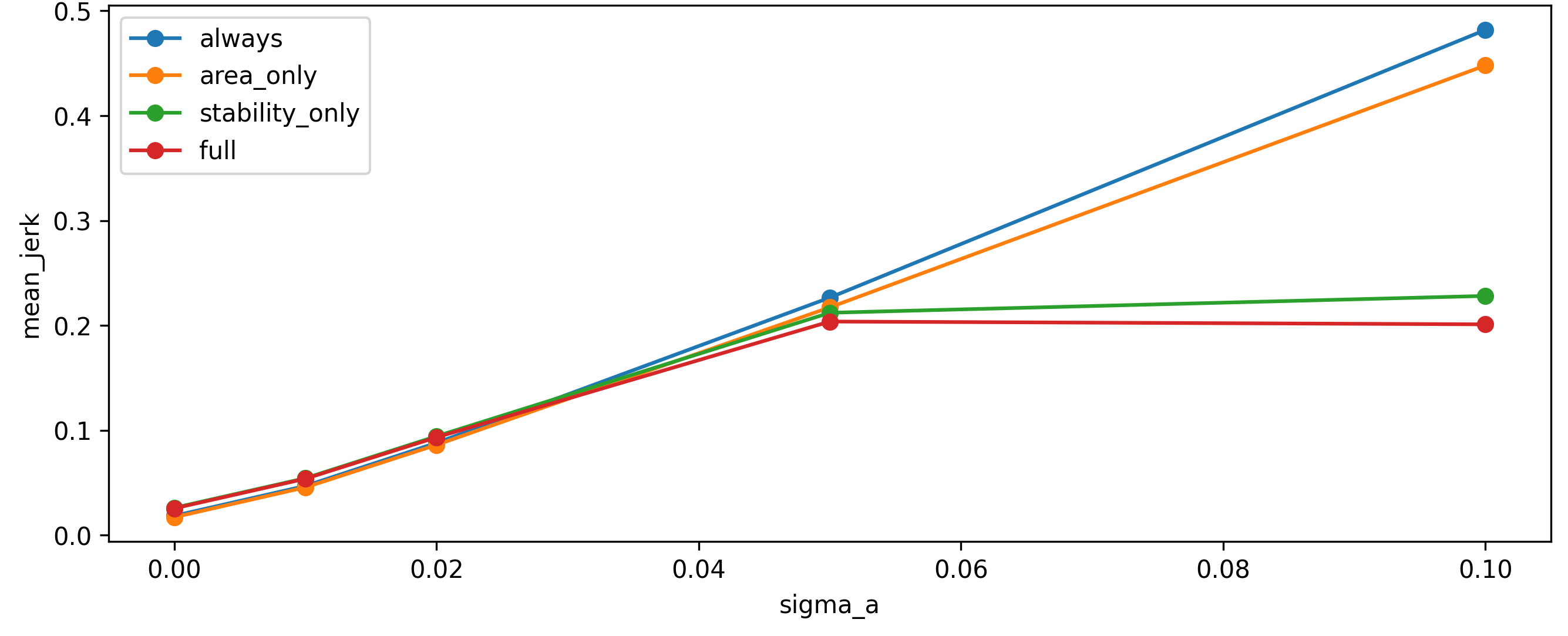}
    \caption{Baseline freeze-HOLD policy: mean jerk versus area perturbation magnitude. The full gate provides the lowest jerk at high area noise, but this is accompanied by reduced availability and increased issued-stream error.}
    \label{fig:orig_area_jerk}
\end{figure}

Figure~\ref{fig:orig_area_mae_exec} shows that stronger gating again improves executed-only MAE. At \(\sigma_a=0.10\), always-execute and area-only remain near 0.228, whereas stability-only and full gating reduce executed-only MAE to approximately 0.170 and 0.166, respectively. Figure~\ref{fig:orig_area_jerk} further shows that the full gate achieves the lowest jerk at high area noise.

\begin{figure}[!h]
    \centering
    \includegraphics[width=0.95\columnwidth]{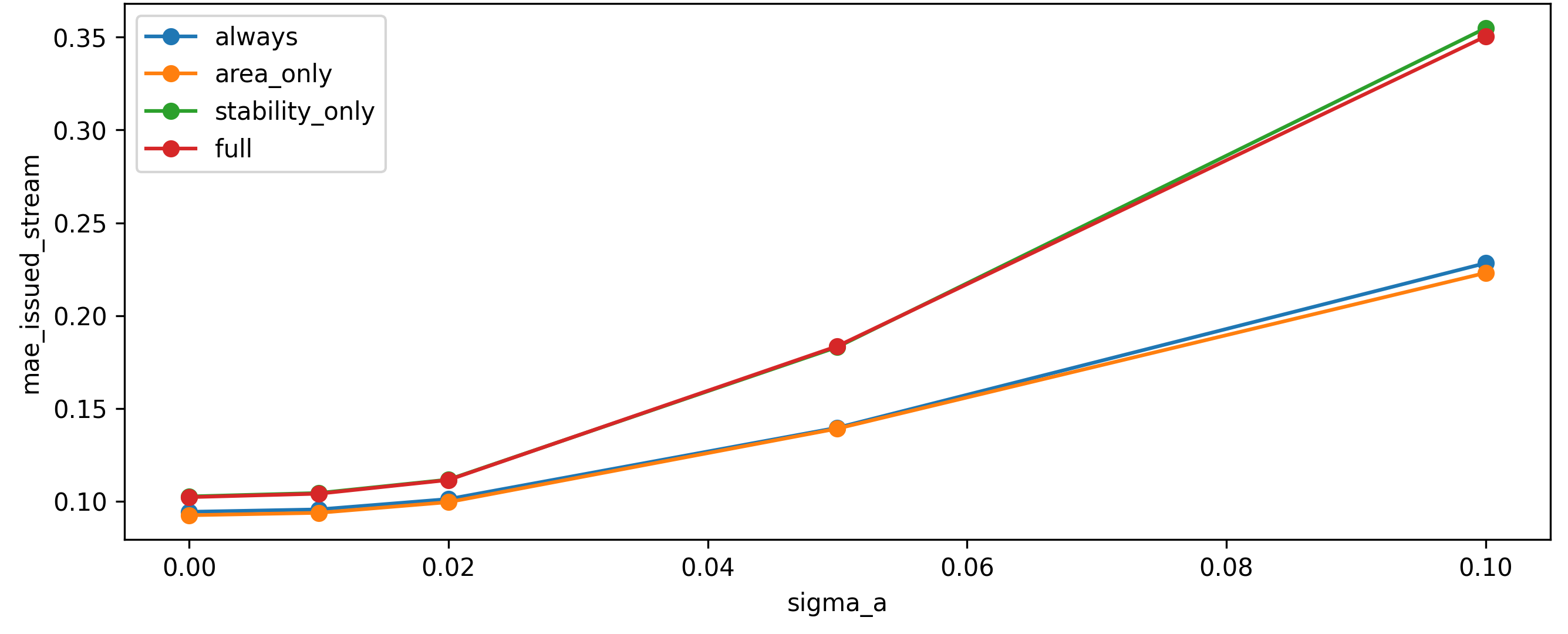}
    \caption{Baseline freeze-HOLD policy: issued-stream MAE versus area perturbation magnitude. Sparse execution degrades the final command stream under the baseline freeze-HOLD policy.}
    \label{fig:orig_area_mae_issued}
\end{figure}

However, the issued-stream MAE in Figure~\ref{fig:orig_area_mae_issued} shows that freeze-HOLD policy still degrades command-level accuracy once execution becomes sparse. At \(\sigma_a=0.10\), stability-only and full gating produce issued-stream MAE values of 0.3549 and 0.3504, both worse than always-execute and area-only. 

Table~\ref{tab:area_high_noise_freeze} summarizes this trade-off. Compared with the centroid-noise case, the degradation is less dramatic, but the same mechanism is present: selective gating improves the executed subset while the baseline freeze-HOLD policy degrades the final command stream when updates become infrequent.

\begin{table}[h] \caption{Baseline freeze-HOLD policy at highest area perturbation level \((\sigma_a = 0.10)\)} \label{tab:area_high_noise_freeze} \centering \setlength{\tabcolsep}{3.0pt} \small \begin{tabular}{lcccc} \toprule Mode & \makecell{Execution\\Rate (\%)} & \makecell{MAE Exec.\\Only (deg)} & \makecell{MAE\\Issued (deg)} & \makecell{Mean Jerk\\(deg/step$^2$)} \\ \midrule Always-execute & 100.00 & 0.2283 & 0.2283 & 0.4818 \\ Area-only & 91.03 & 0.2271 & 0.2230 & 0.4479 \\ Stability-only & 61.77 & 0.1702 & 0.3549 & 0.2281 \\ Full gate & 54.03 & 0.1657 & 0.3504 & 0.2011 \\ \bottomrule \end{tabular} \end{table}

\begin{table*}[!h] \caption{Clean-sequence comparison under the updated bounded-blend HOLD policy} \label{tab:clean_results_blend} \centering \begin{tabular}{lcccc} \hline Mode & Execution Rate (\%) & MAE Executed (deg) & MAE Issued (deg) & Mean Jerk (deg/step$^2$) \\ \hline Always-execute & 100.00 & 0.094285 & 0.094285 & 0.018487 \\ Area-only & 98.78 & 0.091493 & 0.092586 & 0.016980 \\ Stability-only & 97.24 & 0.089771 & 0.099964 & 0.022653 \\ Full gate & 96.35 & 0.088549 & 0.099837 & 0.022432 \\ \hline \end{tabular} \end{table*}

\subsection{Updated Bounded-Blend HOLD Policy}

\subsubsection{Clean-Sequence Behavior}

The updated bounded-blend HOLD policy was evaluated under the same predictor, threshold pair, and gate decisions as the baseline freeze-HOLD policy.

The differences therefore appear only in the issued-command behavior. Table~\ref{tab:clean_results_blend} shows that the bounded-blend update preserves the always-execute baseline exactly and yields small but meaningful improvements for the more selective gates. In particular, issued-stream MAE decreases from 0.102562 to 0.099964 for stability-only and from 0.102184 to 0.099837 for the full gate. Mean jerk also decreases from 0.026114 to 0.022653 for stability-only and from 0.025597 to 0.022432 for the full gate. Thus, the bounded-blend update does not degrade clean-sequence behavior and slightly improves the final command stream even in the absence of perturbation.

\subsubsection{Response to Centroid Perturbation}

Under centroid perturbation, the executed-fraction and executed-only MAE curves remain identical to those of the baseline freeze-HOLD study, confirming that the bounded-blend update does not change the gate behavior itself.

\begin{figure}[h]
    \centering
    \includegraphics[width=\columnwidth]{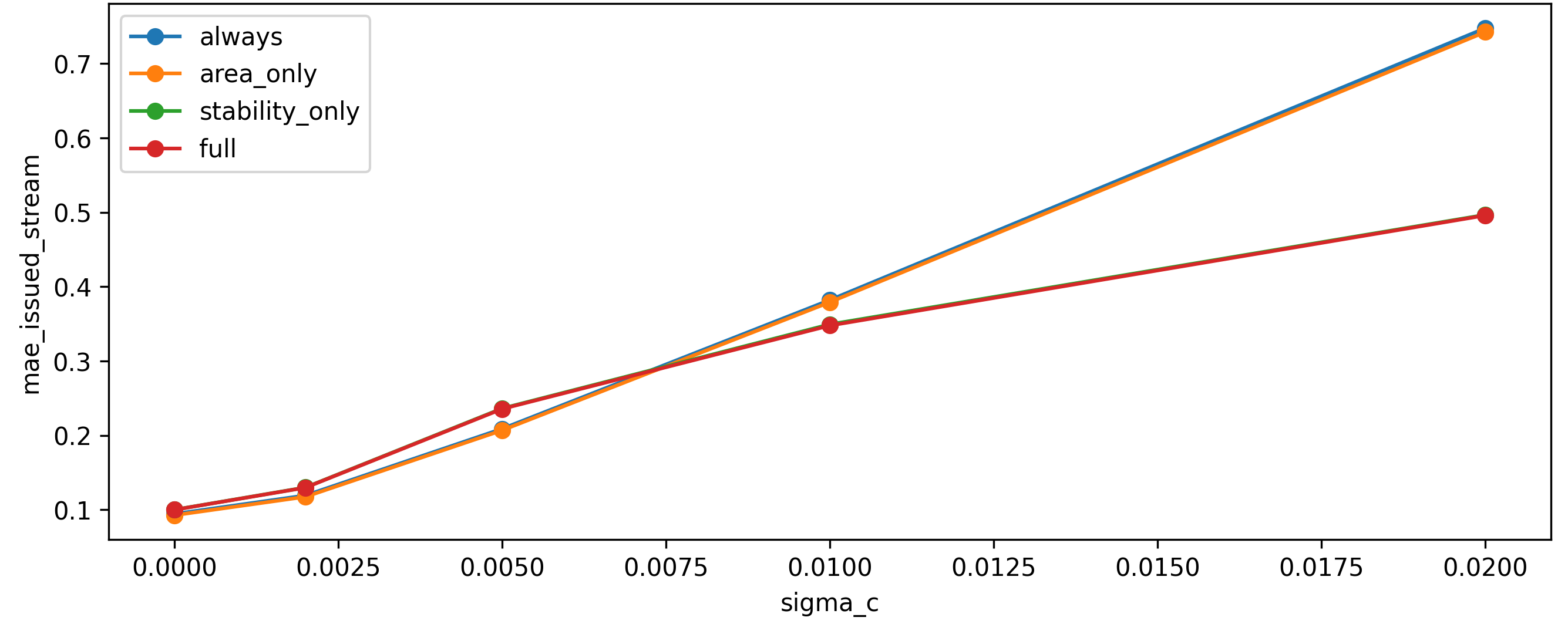}
    \caption{Updated bounded-blend HOLD policy: issued-stream MAE versus centroid perturbation magnitude. The bounded-blend fallback removes the severe stale-command growth observed under the baseline freeze-HOLD policy.}
    \label{fig:blend_centroid_mae_issued}
\end{figure}

The crucial difference appears in the issued-stream MAE. As shown in Figure~\ref{fig:blend_centroid_mae_issued}, the bounded-blend fallback substantially reduces the command-level degradation observed under the freeze-HOLD. At \(\sigma_c=0.01\), issued-stream MAE decreases from approximately 0.75 to approximately 0.35 for both stability-only and full gating. At the highest perturbation level, \(\sigma_c=0.02\), issued-stream MAE decreases from 3.2176 to 0.4966 for stability-only and from 3.2152 to 0.4959 for the full gate. These values are summarized in Table~\ref{tab:high_noise_comparison}.

\begin{figure}[h]
    \centering
    \includegraphics[width=\columnwidth]{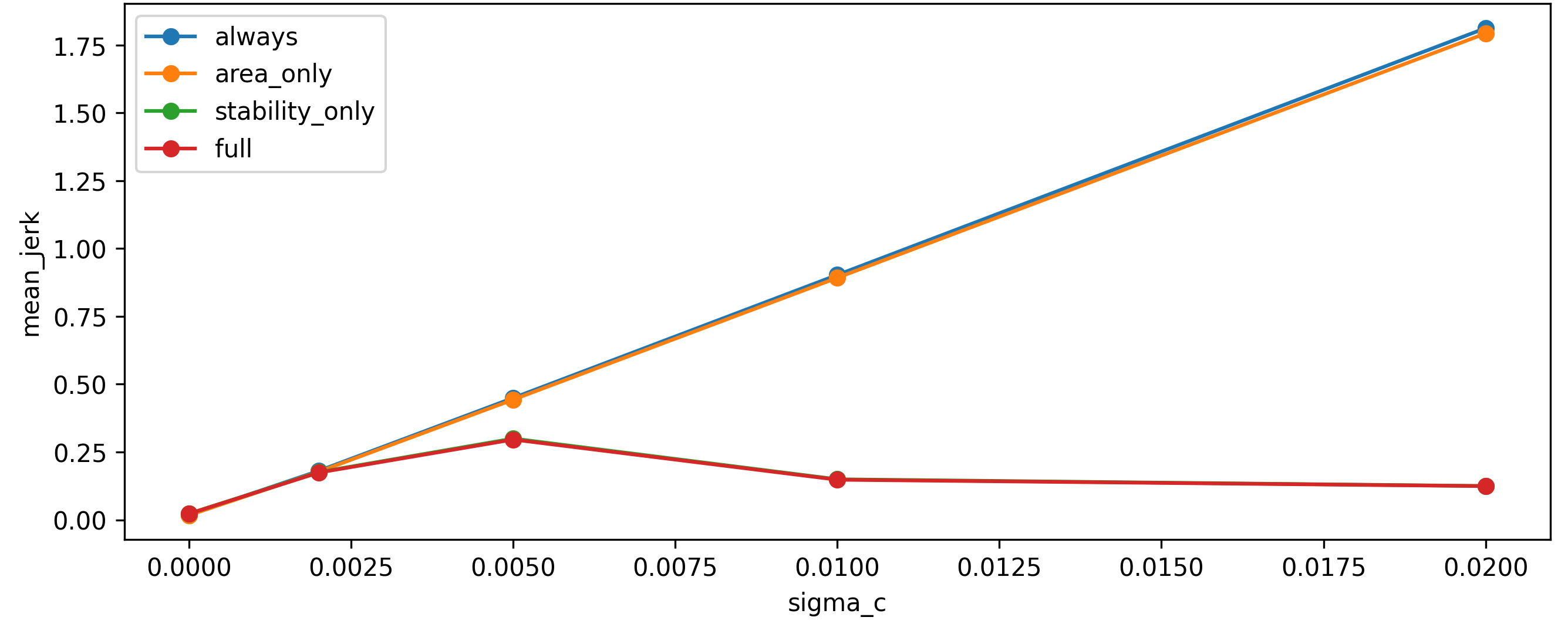}
    \caption{Updated bounded-blend HOLD policy: mean jerk versus centroid perturbation magnitude. Jerk increases modestly relative to the baseline freeze-HOLD policy, reflecting conservative adaptation rather than indefinite freezing.}
    \label{fig:blend_centroid_jerk}
\end{figure}

Figure~\ref{fig:blend_centroid_jerk} shows that this improvement is accompanied by a modest increase in jerk at the highest centroid perturbation level. For the full gate, mean jerk increases from 0.0805 to 0.1244, and for stability-only from 0.0812 to 0.1247. This increase is expected, because the command is no longer frozen but is allowed to adapt conservatively during low-confidence intervals. The trade-off is clearly favorable: a moderate increase in motion activity prevents the severe stale-command accumulation observed under the baseline freeze-HOLD policy.

\subsubsection{Response to Area Perturbation}

The same pattern appears under area perturbation. Executed fractions and executed-only MAE values remain unchanged from the baseline freeze-HOLD case, indicating that the gate continues to make the same execution decisions. However, the issued-stream behavior improves under the bounded-blend fallback.

Figure~\ref{fig:blend_area_mae_issued} shows that the issued-stream MAE decreases across the more selective gates. At \(\sigma_a=0.10\), issued-stream MAE decreases from 0.3549 to 0.2531 for stability-only and from 0.3504 to 0.2477 for the full gate. Unlike the centroid-noise case, smoothness also improves slightly:
\begin{figure}[h]
    \centering
    \includegraphics[width=\columnwidth]{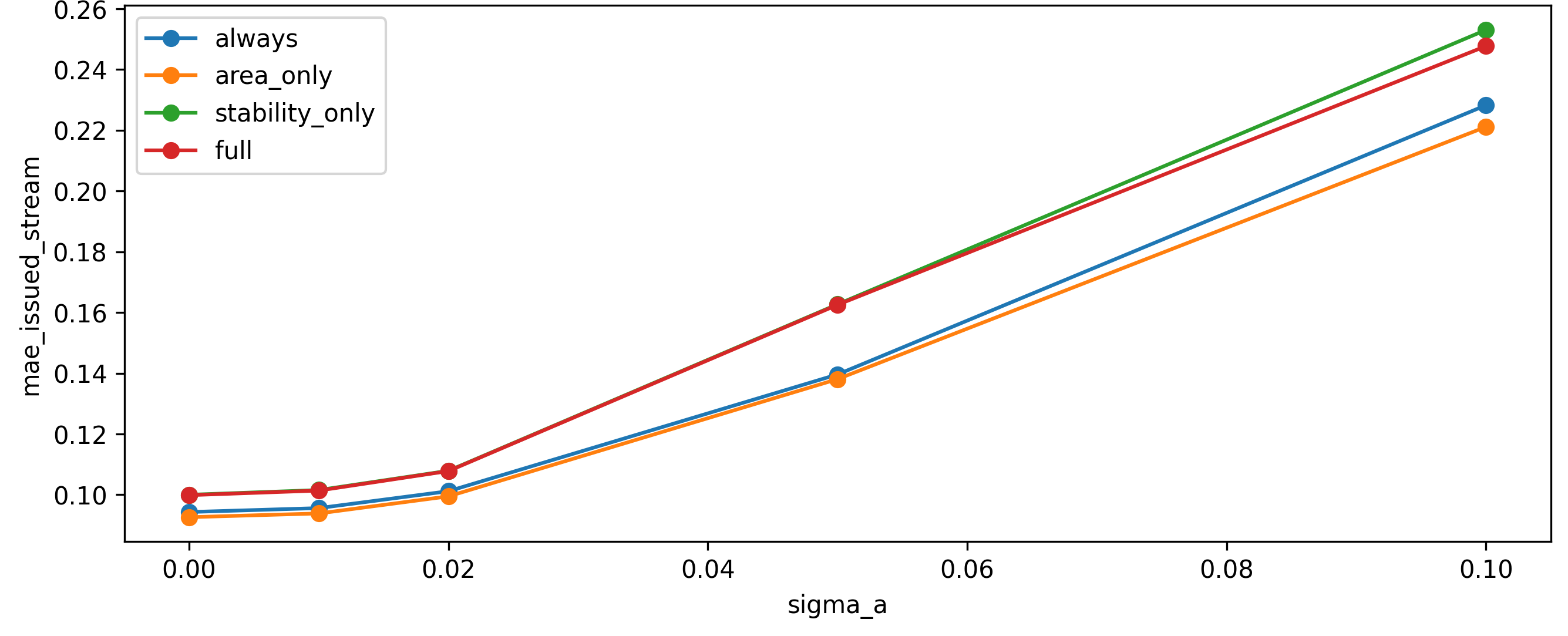}
    \caption{Updated bounded-blend HOLD policy: issued-stream MAE versus area perturbation magnitude. The bounded-blend fallback improves command-level accuracy compared with the baseline freeze-HOLD policy.}
    \label{fig:blend_area_mae_issued}
\end{figure}

\begin{figure}[h]
    \centering
    \includegraphics[width=\columnwidth]{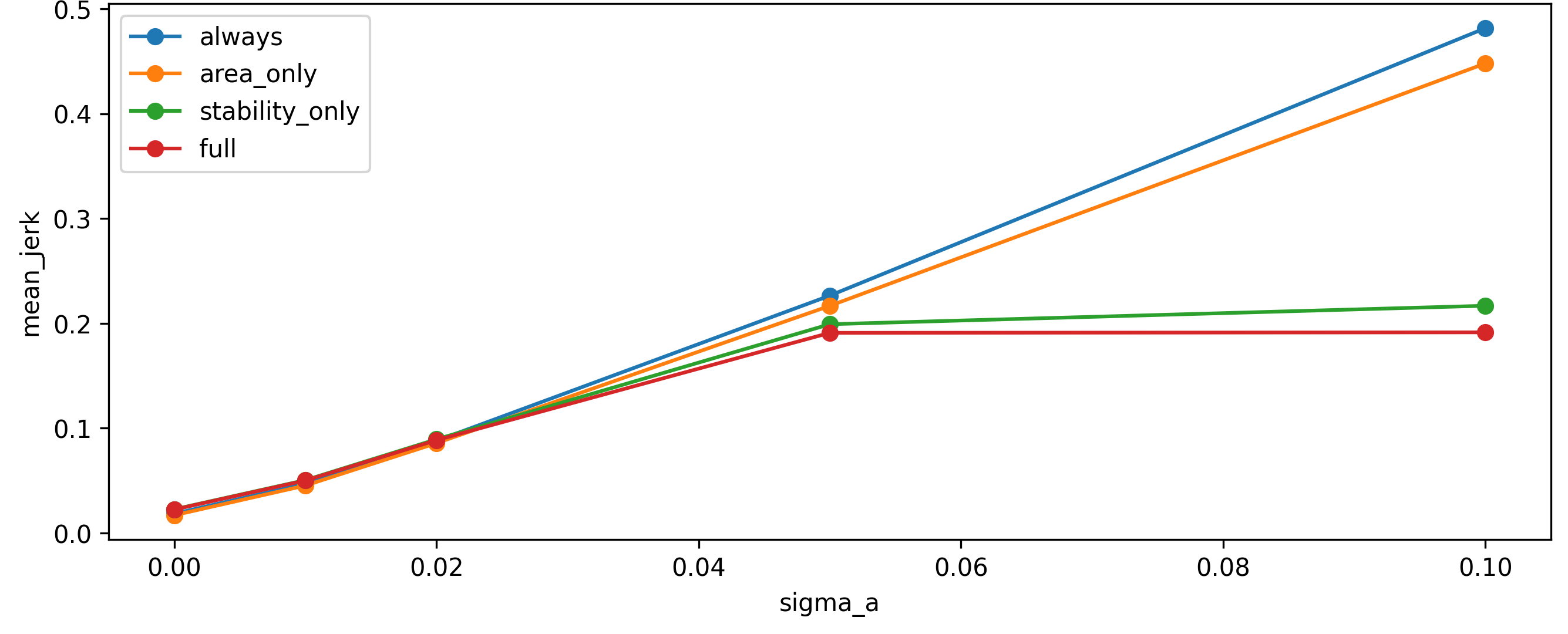}
    \caption{Updated bounded-blend HOLD policy: mean jerk versus area perturbation magnitude. The updated fallback slightly improves smoothness at high area perturbation.}
    \label{fig:blend_area_jerk}
\end{figure}

Figure~\ref{fig:blend_area_jerk} shows that mean jerk decreases from 0.2281 to 0.2169 for stability-only and from 0.2011 to 0.1915 for the full gate. Therefore, under area perturbation, the bounded-blend update provides a more favorable trade-off in both command accuracy and smoothness.

\subsection{Comparative Discussion}

\begin{table*}[h] \caption{Comparison between the baseline freeze-HOLD and bounded-blend HOLD policies at the highest perturbation levels} \label{tab:high_noise_comparison} \centering \begin{tabular}{llcccc} \hline Condition & Mode & Execution Rate (\%) & MAE Exec. Only (deg) & MAE Issued Freeze (deg) & MAE Issued Blend (deg) \\ \hline \multirow{2}{*}{\(\sigma_c = 0.02\)} & Stability-only & 1.66 & 0.4445 & 3.2177 & 0.4966 \\ & Full gate & 1.63 & 0.4397 & 3.2153 & 0.4959 \\ \hline \multirow{2}{*}{\(\sigma_a = 0.10\)} & Stability-only & 61.77 & 0.1702 & 0.3549 & 0.2531 \\ & Full gate & 54.03 & 0.1657 & 0.3504 & 0.2477 \\ \hline \end{tabular} \end{table*}

The comparison between the baseline freeze-HOLD and updated HOLD policies reveals that the confidence gate and the fallback rule play distinct but tightly coupled roles. The gate determines how selectively the system accepts new predictions, whereas the fallback rule determines how the issued command evolves once execution becomes sparse. In the present study, the threshold pair, executed fractions, and executed-only accuracies remain unchanged between the two HOLD policies. Therefore, the observed differences in command-level behavior can be attributed directly to the fallback update mechanism.

The baseline freeze-HOLD reveals an important limitation of relying on confidence gating alone. Under strong centroid perturbation, the gate successfully rejects unreliable predictions and preserves a cleaner executed subset, but indefinite freezing causes the issued command to become stale once updates become too rare. This failure mode is especially clear at \(\sigma_c=0.02\), where stability-only and full gating reduce executed-only MAE to about 0.44 while issued-stream MAE rises above 3.21.

The bounded-blend update addresses this limitation while preserving the same selective execution behavior. Because the issued command is allowed to adapt conservatively during HOLD, the severe stale-command growth is removed. At \(\sigma_c=0.02\), issued-stream MAE decreases from 3.2176 to 0.4966 for stability-only and from 3.2152 to 0.4959 for the full gate. Under area perturbation, the same update also improves the command-level trade-off, reducing issued-stream MAE and slightly improving jerk. The main quantitative comparison at the highest perturbation levels is summarized in Table~\ref{tab:high_noise_comparison}.

These results show that confidence gating alone is not sufficient for reliable perception-to-action behavior once execution becomes sparse. The fallback policy is not merely an implementation detail; it is part of the decision framework itself. In the present study, the bounded-blend HOLD policy provides a substantially better command-level trade-off than the baseline freeze-HOLD policy, making the overall confidence-gated framework more robust to stale-command degradation while preserving conservative execution under uncertainty.


\section{Conclusion}

This paper studied decision-level command issuance for vision-based UAV--UGV heading alignment using a fixed inherited heading predictor. Instead of redesigning the perception model, the work focused on how predicted headings should be translated into issued commands when perceptual reliability varies over time. To address this problem, the paper introduced a confidence-gated framework based on two lightweight reliability proxies: a visibility-related proxy derived from bounding-box area and a stability-related proxy derived from short-window variation in the predicted heading sequence.

The results showed that confidence gating creates a clear trade-off among execution availability, executed-frame accuracy, issued-command accuracy, and smoothness. Under degraded perception, the gate became more selective and preserved a cleaner executed subset. However, the experiments also showed that the baseline freeze-HOLD policy can cause severe issued-stream degradation when execution becomes too sparse. Under the same predictor, thresholds, and gate decisions, the bounded-blend HOLD policy substantially reduced stale-command accumulation and produced a more favorable command-level trade-off.

Overall, the findings show that reliable perception-driven autonomy depends not only on deciding when a predicted command should be executed, but also on designing how the issued command should evolve when execution is withheld. Future work can extend this framework by incorporating additional reliability proxies, learning or adapting thresholds online, and evaluating the method in longer closed-loop experiments with more diverse operating conditions.



\bibliographystyle{IEEEtran}
\bibliography{ref}

 
\vspace{11pt}




\vfill

\end{document}